\documentclass[10pt,twocolumn,letterpaper]{article}

\usepackage{cvpr}
\usepackage{times}
\usepackage{epsfig}
\usepackage{amsmath}
\usepackage{booktabs}       
\usepackage{amssymb}
\usepackage[dvipsnames]{xcolor}
\usepackage{pifont}
\usepackage{comment}
\usepackage{graphicx}

\newcommand{\cmt}[1]{\iffalse #1 \fi}

\usepackage[pagebackref=true,breaklinks=true,letterpaper=true,colorlinks,bookmarks=false]{hyperref}
\usepackage{cleveref}

\Crefname{equation}{Eq.}{Eqs.}
\Crefname{figure}{Fig.}{Figs.}
\Crefname{tabular}{Tab.}{Tabs.}
\Crefname{table}{Tab.}{Tabs.}

\def\cup{}

\def\ceq{}

\cvprfinalcopy 

\def\cvprPaperID{****} 

\def\eg{\emph{e.g.}}

\def\etal{\emph{et al.}}

\def\dash{-\phantom{.0}}
\def\ndash{-}
\def\amp{\&} 

\def\ab{\extracolsep{\fill}}

\begin{document}

\title{Colorization as a Proxy Task for Visual Understanding}

\author{Gustav Larsson\\
University of Chicago\\
{\tt\small larsson@cs.uchicago.edu}
\and
Michael Maire\\
TTI Chicago\\
{\tt\small mmaire@ttic.edu}
\and
Gregory Shakhnarovich\\
TTI Chicago\\
{\tt\small greg@ttic.edu}
}

\maketitle

\begin{abstract}
We investigate and improve self-supervision as a drop-in replacement for ImageNet pretraining, focusing on automatic colorization as the proxy task.  Self-supervised training has been shown to be more promising for utilizing unlabeled data than other, traditional unsupervised learning methods. We build on this success and evaluate the ability of our self-supervised network in several contexts. On VOC segmentation and classification tasks, we present results that are state-of-the-art among methods not using ImageNet labels for pretraining representations.

Moreover, we present the first in-depth analysis of self-supervision via colorization, concluding that formulation of the loss, training details and network architecture play important roles in its effectiveness. This investigation is further expanded by revisiting the ImageNet pretraining paradigm, asking questions such as: How much training data is needed? How many labels are needed? How much do features change when fine-tuned? We relate these questions back to self-supervision by showing that colorization provides a similarly powerful supervisory signal as various flavors of ImageNet pretraining.

\end{abstract}


\begin{figure}
\centering
\includegraphics[width=1.02\linewidth]{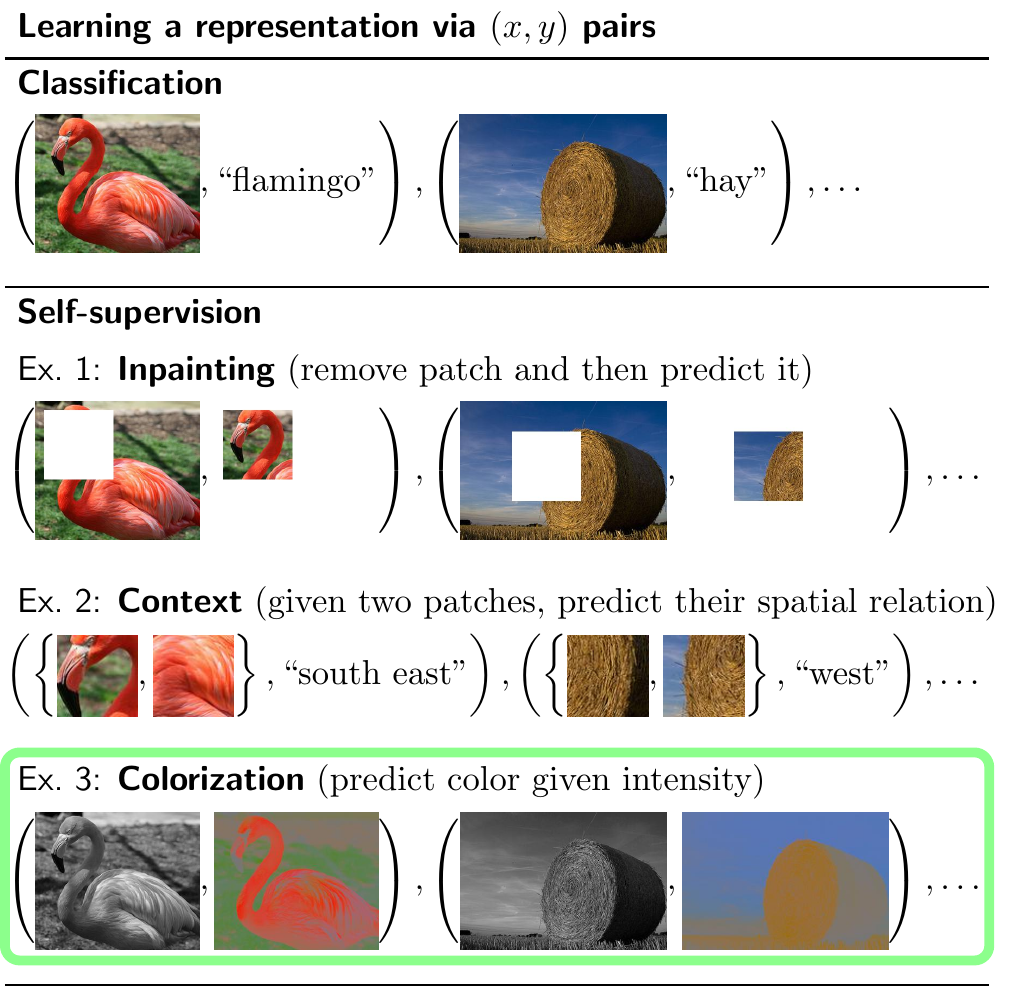}
\caption{
    Using a representation that was originally trained for classification on $(x, y)$
    pairs to initialize a network has become standard practice
    in computer vision. Self-supervision is a family of alternative
    pretraining methods that do not require any labeled data, since labels are
    ``manufactured" through unlabeled data. We focus on colorization,
    where an image is split into its intensity and color components, the former
    predicting the latter.
}
\label{fig:overview}
\end{figure}

\section{Introduction}

The success of deep feed-forward networks is rooted in their ability to scale
up with more training data. The availability of more data can
generally afford an increase in model complexity. However, this
need for expensive, tedious and error-prone human annotation is
severely limiting, reducing our ability to build models for new
domains, and for domains in which annotations are particularly
expensive (e.g., image segmentation). At the same time, we have access
to enormous amounts of unlabeled visual data, which is essentially
free. This work is an attempt to improve means of leveraging
this abundance. We manage to bring it one step closer to the results
of using labeled data, but the eventual long term goal of self-supervision
may be to supplant supervised pretraining completely.

Alternatives to supervised training that do not need labeled data have seen
limited success. Unsupervised learning methods, such as compressed embeddings
trained by minimizing reconstruction error, have seen more success in image
synthesis~\cite{kingma2014auto}\cmt{ and compression}, than for representation
learning. Semi-supervised learning, jointly training a supervised and an
unsupervised loss, offers a middle
ground~\cite{grandvalet2004semi,sajjadi2016semi}. However, recent works tend to
prefer a sequential combination instead (unsupervised pretraining, supervised
fine-tuning)~\cite{doersch2015unsupervised,donahue2016adversarial}, possibly
because it prevents the unsupervised loss from being disruptive in the
late stages of training. A related endeavor
to unsupervised learning is developing models that work with weaker forms of
supervision~\cite{bearman2016point,xu2015weak}. This reduces the human burden
only somewhat and pays a price in model performance.

Recently, {\em self-supervision} has emerged as a new flavor of unsupervised
learning~\cite{doersch2015unsupervised,wang2015unsupervised}. The key
observation is that perhaps part of the benefit of labeled data is that it
leads to using a discriminative loss.  This type of loss may be better
suited for representation learning than, for instance, a reconstruction or
likelihood-based loss. Self-supervision is a way to use a discriminative loss
on unlabeled data by partitioning each input sample in two, predicting the
parts' association. We focus on self-supervised
colorization~\cite{larsson2016learning,zhang2016colorful}, where each image is
split into its intensity and its color, using the former to predict the latter.

Our main contributions to self-supervision are:

\begin{itemize}
    \item State-of-the-art results on VOC 2007 Classification and VOC 2012
        Segmentation, among methods that do not use ImageNet labels.
    \item The first in-depth analysis of self-supervision via colorization. We
        study the impact of loss, network architecture and training details,
        showing that there are many important aspects that influence results.
    \item An empirical study of various formulations of ImageNet pretraining
        and how they compare to self-supervision.
\end{itemize}

\section{Related work}

In our work on replacing classification-based pretraining for downstream supervised tasks, the first thing to consider is clever network initializations.
Networks that are initialized to promote uniform scale of activations across
layers, converge more easily and
faster~\cite{glorot2010understanding,he2015msra}. The uniform scale however is
only statistically predicted given broad data assumptions, so this idea can be
taken one step further by looking at the activations of actual data and
normalizing~\cite{mishkin2015all}. Using some training data to initialize
weights blurs the line between initialization and unsupervised
pretraining. For instance, using layer-wise $k$-means
clustering~\cite{coates2010analysis,krahenbuhl2016datadriven} should be
considered unsupervised pretraining, even though it may be a particularly fast
one.

Unsupervised pretraining can be used to facilitate optimization or to expose
the network to orders of magnitude larger unlabeled data. The former was once a
popular motivation, but fell out of favor as it was made unnecessary by
improved training techniques~(\eg\ introduction of non-saturating
activations~\cite{nair2010rectified}, better
initialization~\cite{glorot2010understanding} and training
algorithms~\cite{qian1999momentum,kingma2015adam}). The second motivation of
leveraging more data, which can also be realized as semi-supervised training,
is an open problem with current best methods rarely used in competitive vision
systems.

Recent methods on self-supervised feature learning have seen several
incarnations, broadly divided into methods that exploit temporal
or spatial structure in natural visual data:

\textbf{Temporal.}
There have been a wide variety of methods that use the correlation between
adjacent video frames as a learning signal. One way
is to try to predict future frames, which is an analogous task to language
modeling and often uses similar techniques based on RNNs and
LSTMs~\cite{srivastava2015video,ranzato2014video}. It is also possible to train
an embedding where temporally close frames are considered similar (using either
pairs~\cite{mobahi2009deep,isola2015learning,jayaraman2015slow} or
triplets~\cite{wang2015unsupervised}). Another method that uses a triplet loss
presents three frames and tries to predict if they are correctly
ordered~\cite{misra2016unsupervised}. Pathak~\etal~\cite{pathak2016move} learn
general-purpose representation by predicting saliency based on optical flow.
Owens~\etal~\cite{owens2016ambient},
somewhat breaking from the temporal category, operate on a single video frame
to predict a statistical summary of the audio from the entire clip. The first
video-based self-supervision methods were based on Independent Component
Analysis (ICA)~\cite{van1998independent,hurri2003simple}. Recent
follow-up work generalizes this to a nonlinear
setting~\cite{hyvarinen2016unsupervised}.

\textbf{Spatial.} Methods that operate on single-frame input typically use the
spatial dimensions to divide samples for self-supervision. Given a pair of
patches from an image, Doerch~\etal~\cite{doersch2015unsupervised} train
representations by predicting which of eight possible spatial compositions the
two patches have. Noroozi \& Favaro~\cite{noroozi2016jigsaw} take this further and learns a representation by solving a 3-by-3 jigsaw
puzzle. The task of inpainting (remove some pixels, then predict
them) is utilized for representation learning by
Pathak~\etal~\cite{pathakCVPR16context}.  There has also been work on using
bi-directional Generative Adversarial Networks (BiGAN) to learn
representations~\cite{donahue2016adversarial,dumoulin2016adversarially}. This is not what we typically
regard as self-supervision, but it does similarly pose a supervised learning
task (\emph{real} vs.\ \emph{synthetic}) on unlabeled data to drive
representation learning.

\textbf{Colorization.} Lastly there is
colorization~\cite{larsson2016learning,zhang2016colorful,zhang2017split}.
Broadly speaking, the two
previous categories split input samples along a spatio-temporal line, either
predicting one given the other or predicting the line itself. Automatic
colorization departs from this as it asks to predict color over the same pixel
as its center of input, without discarding any spatial information. We
speculate that this may make it more suitable to
tasks of similar nature, such as semantic segmentation; we demonstrate strong
results on this benchmark.

Representation learning via colorization was first presented as part of two
automatic colorization papers~\cite{larsson2016learning,zhang2016colorful}.
Zhang~\etal~\cite{zhang2016colorful} present results across all PASCAL tasks
and show colorization as a front-runner of self-supervision.
However, like most self-supervision papers, it is restricted to AlexNet and
thus shows modest results compared to recent supervised methods.
Larsson~\etal~\cite{larsson2016learning} present state-of-the-art results on
PASCAL VOC semantic segmentation, which we improve by almost 10 points from
50.2\% to 60.0\% mIU. Both papers present the results with little analysis or
investigation.

\begin{figure}[t]
    \centering
    \includegraphics{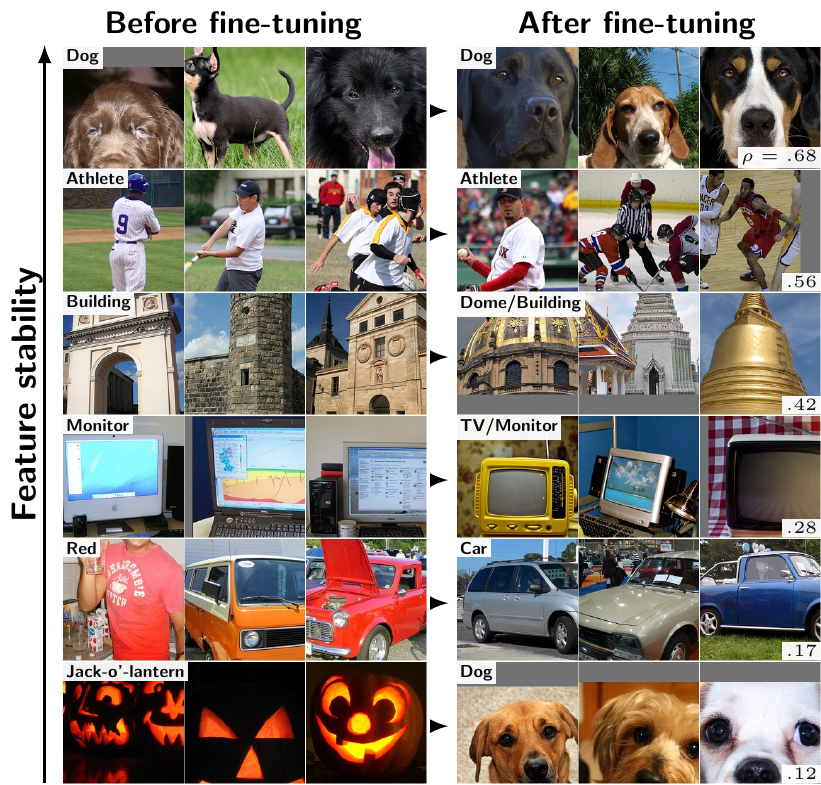}
    \caption{
        \textbf{Feature reuse/repurpose.} The left column visualizes top
        activations from the colorization network (same as in~\Cref{fig:viz}).
        The right column visualizes the corresponding feature after the network
        has been fine-tuned for semantic segmentation. Features are either
        re-used as is (top), specialized (middle), or scrapped and replaced (bottom).
        See~\Cref{fig:corr-after-finetune} for a quantitative study.
    }
    \label{fig:viz-after-finetune}
\end{figure}

\begin{figure}
    \centering
    \includegraphics[width=1.0\linewidth]{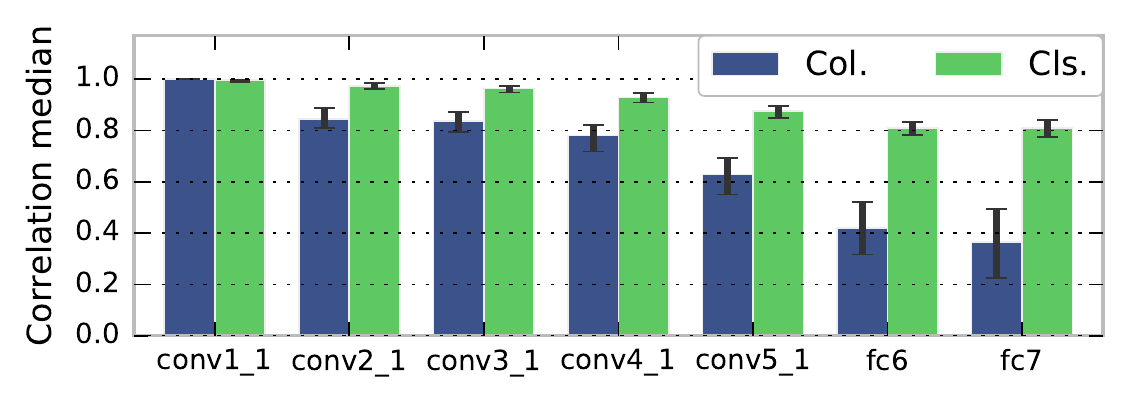}
    \caption{
        \textbf{Feature shift.} The correlation between feature activations for
        layers of VGG-16 before and after fine-tuning for semantic segmentation. The
        bar heights indicate median correlation and error bars indicate
        interquartile range. See~\Cref{fig:viz-after-finetune} for qualitative
        examples.
    }
    \label{fig:corr-after-finetune}
\end{figure}

\section{Colorization as the target task}

Training an automatic colorizer for the purpose of being able to convert
grayscale photos to color is an active area of
research~\cite{larsson2016learning,zhang2016colorful,iizuka2016color}. Recent
methods train deep convolutional neural networks to predict
color~\cite{iizuka2016color} or distributions over
color~\cite{larsson2016learning,zhang2016colorful}. The latter approach is
followed by instantiating a color from the histogram prediction in order to
produce a final result. For optimal colorization results, these networks are
initialized with a classification-based network, in order to leverage its
high-level features and thus better predict color. In this section we describe
how to train colorization, revisiting some of the design decisions that were
made with the goal of producing aesthetic color images and instead consider
their impact on learning representations.

\subsection{Training}

Our experimental setup borrows heavily from
Larsson~\etal~\cite{larsson2016learning}, using Caffe~\cite{jia2014caffe} and their public source code release for
training the colorization network. For downstream tasks, we use
TensorFlow~\cite{tensorflow2015-whitepaper} and provide testing code as well as trained models.\footnote{\url{https://github.com/gustavla/self-supervision}}

\textbf{Loss.} We consider both a regression loss for
L*a*b color
values~\cite{larsson2016learning,zhang2016colorful,iizuka2016color}, as well as
a KL divergence loss for hue/chroma histograms~\cite{larsson2016learning}. For
the latter, the histograms are computed from a 7-by-7 window around each target
pixel and placed into 32 bins for hue and 32 bins for chroma. We evaluate
their ability to learn representations, disregarding their ability to do
colorization. In our comparison, we make sure that the losses are scaled
similarly, so that their effective learning rates are as close as possible.

\textbf{Hypercolumn.} The networks use
hypercolumns~\cite{MYP:ACCV:2014,mostajabi2015feedforward,
hariharan2015hypercolumns} with sparse training~\cite{larsson2016learning}.
This means that for each image, only a small sample of hypercolumns are
computed. This reduces memory requirements and allows us to train on larger
images. Note that hypercolumns can be used for colorization pretraining, as
well as for segmentation as a downstream task.  Since we have reasons to
believe that hypercolumn training may disrupt residual training, we do not
train our ResNet colorizer with hypercolumns.

\textbf{Dataset.} We train on 3.7M \emph{unlabeled} images by combining 1.3M from
ImageNet~\cite{imagenet} and 2.4M from Places205~\cite{zhou2014learning}. The
dataset contains some grayscale images, but we do not make an effort to sort
them out, since there is no way to tell a legitimately achromatic image from a
desaturated one.

\textbf{Training.} All training is done with standard Stochastic Gradient
Descent with momentum set to 0.9. The colorization network is initialized with
Xavier initialization~\cite{glorot2010understanding} and trained with batch
normalization without re-biasing or re-scaling
parameters~\cite{ioffe2015batch}. Each time an image is processed, it is
randomly mirrored and the image is randomly scaled such that the shortest side
is between 352 and 600. Finally, a 352-by-352 patch is extracted and
desaturated and then fed through the network. In our comparative studies, we
train using a colorization loss for 3 epochs (spending 2 epochs on the initial
learning rate). In our longer running experiments, we trained for about 10
epochs. For our best ResNet model, we trained significantly longer (35 epochs),
although on smaller inputs (224-by-224); we found large input sizes to be
more important during downstream training.

\section{Colorization as a proxy task}

Shifting our focus to using a colorization network purely for its visual
representations, we describe how it can help improve results on classification
and segmentation.

\subsection{Training}
The downstream task is trained by initializing weights from the
colorization-from-scratch network. Some key considerations follow:

\textbf{Early stopping.} Training on a small sample size is prone to
overfitting. We have found that the most effective method of preventing this is
carefully cross validating the learning rate schedule. Models that initialize
differently (random, colorization, classification), need very different early
stopping schedules. Finding a method that works well in all these settings was
key to our study. We split
the training data 90/10 and only train on the 90\%; the rest is used to monitor
overfitting. Each time the 10\% validation score (not surrogate loss) stops
improving, the learning rate is dropped. After this is done twice, the training
is concluded. For our most competitive experiments (\Cref{tab:comparison}), we
then re-train using 100\% of the data with the cross-validated learning rate
schedule fixed.

\textbf{Receptive field.} Previous work on semantic segmentation has shown the
importance of large receptive
fields~\cite{mostajabi2015feedforward,yu2015multiscale}. One way of
accomplishing this is by using dilated
convolutions~\cite{yu2015multiscale,wu2016resnet}, however this redefines the
interpretation of filters and thus requires re-training.  Instead, we add two
additional blocks (2-by-2 max pooling of stride 2, 3-by-3 convolution with 1,024 features)
at the top of the network, each expanding the receptive field with 160 pixels
per block. We train on large input images (448-by-448) in order to fully appreciate
the enlarged receptive field.

\textbf{Hypercolumn.} Note that using a hypercolumn when the downstream task is
semantic segmentation is a separate design choice that does not need to be
coupled with the use of hypercolumns during colorization pretraining. In either
case, the post-hypercolumn parameter weights are never re-used. For ResNet, we
use a subset of the full hypercolumn.\footnote{ResNet-152 hypercolumn:
\texttt{conv1}, \texttt{res2\{a,b,c\}}, \texttt{res3b\{1,4,7\}},
\texttt{res4b\{5,10,15,20,25,30,35\}}, \texttt{res5c}}

\textbf{Batch normalization.} The models trained from scratch use
parameter-free batch normalization. However, for downstream training, we absorb
the mean and variance into the weights and biases and train without batch
normalization (with the exception of ResNet, where in our experience it helps).
For networks that were not trained with batch normalization and are not
well-balanced in scale across layers (\eg~ImageNet pretrained VGG-16), we
re-balance the network so that each layer's activation has unit
variance~\cite{larsson2016learning}.

\begin{table}
\begin{center}
    \begin{tabular*}{\linewidth}{@{\ab}ll|rr}
        \toprule
        Initialization                                                          & Architecture    & Class.                  & Seg.                           \\
                                                                                &                 & {\footnotesize \%mAP}   & {\footnotesize \%mIU}          \\ \midrule
        ImageNet \quad\quad\quad{\scriptsize (+FoV)}  \scriptsize               & VGG-16          & 86.9\cmt{a230}          & 69.5                           \\ \toprule
        Random (ours)                                                           & AlexNet         & 46.2\cmt{a231}          & 23.5\cmt{a228}        \\
        Random~\cite{pathakCVPR16context}                                       & AlexNet         & 53.3                    & 19.8\cmt{H:a116}               \\
        $k$-means~\cite{krahenbuhl2016datadriven,donahue2016adversarial}        & AlexNet         & 56.6                    & 32.6                           \\
        $k$-means~\cite{krahenbuhl2016datadriven}                               & VGG-16          & 56.5                    & \dash                          \\
        $k$-means~\cite{krahenbuhl2016datadriven}                               & GoogLeNet       & 55.0                    & \dash                          \\ \midrule
        Pathak~\etal~\cite{pathakCVPR16context}                                 & AlexNet         & 56.5                    & 29.7                           \\
        Wang \amp\ Gupta~\cite{wang2015unsupervised}                            & AlexNet         & 58.7                    & \dash                          \\
        Donahue~\etal~\cite{donahue2016adversarial}                             & AlexNet         & 60.1                    & 35.2                           \\
        Doersch~\etal~\cite{doersch2015unsupervised,donahue2016adversarial}     & AlexNet         & 65.3                    & \dash                          \\
        Zhang~\etal~(col)~\cite{zhang2016colorful}                              & AlexNet         & 65.6                    & 35.6                           \\
        Zhang~\etal~(s-b)~\cite{zhang2017split}\cmt{$^\dagger$}                 & AlexNet         & 67.1                    & 36.0                           \\
        Noroozi \amp\ Favaro~\cite{noroozi2016jigsaw}                           & Mod. AlexNet    & 68.6                    & \dash                          \\
        Larsson~\etal~\cite{larsson2016learning}                                & VGG-16          & \dash                   & 50.2                           \\ \midrule
        Our method                                                              & AlexNet         & 65.9\cmt{c107d/ds-cls}  & 38.4\cmt{a222}                 \\
         \multicolumn{1}{r}{\scriptsize (+FoV)}                                 & VGG-16          & \textbf{77.2}\cmt{a224} & 56.0\cmt{a138}                 \\
        \multicolumn{1}{r}{\scriptsize (+FoV)}                                  & ResNet-152      & \textbf{77.3}\cmt{a246} & \textbf{60.0}\cmt{a244}        \\
        \bottomrule
    \end{tabular*}
    \caption{
        \small
        \textbf{VOC Comparison.} Comparison with other initialization and
        self-supervision methods on VOC 2007 Classification (test) and VOC 2012
        Segmentation (val). Note that our baseline AlexNet results (38.4\%) are
        also the most competitive among AlexNet models. The use of
        a hypercolumn instead of FCN is partly responsible: running
        Zhang~\etal's colorization model with a hypercolumn yields 36.4\%\cmt{218}, only a
        slight improvement over 35.6\%.
        Switching to ResNet, adding a larger FoV, and training even longer
        yields a significantly higher result at 60.0\%\cmt{a221} mIU. Note, the
        ``+FoV'' only affects the segmentation results. 
        The modified AlexNet
        used by Noroozi \amp{} Favaro has the same number of parameters as
        AlexNet, with a spatial reduction of 2 moved from conv1 to pool5, increasing
        the size of the intermediate activations.  \cmt{($^\dagger$ Concurrent work)}
    }
    \label{tab:comparison}
\end{center}
\end{table}

\textbf{Padding.} For our ImageNet pretraining experiments, we observe that
going from a classification network to a fully convolutional network can
introduce edge effects due to each layer's zero padding. A problem not
exhibited by the original VGG-16, leading us to suspect that it may be due to
the introduction of batch normalization. For the newly trained networks,
activations increase close to the edge, even though the receptive fields
increasingly hang over the edge of the image, reducing the amount of semantic
information. Correcting for this\footnote{We pad with the bias from the
previous layer, instead of with zeros. This is an estimate of the expectation value,
since we use a parameter-free batch normalization with zero mean, leaving
only the bias.} makes activations well-behaved, which was important in order to
appropriately visualize top activations. However, it does not offer a
measurable improvement on downstream tasks, which means the network can
correct for this during the fine-tuning stage.

\textbf{Color.} Since the domain of a colorization network is grayscale, our
downstream experiments operate on grayscale input unless otherwise stated. When
colorization is re-introduced, we convert the grayscale filters in
\texttt{conv1\_1} to RGB (replicate to all three channels, divide by three) and
let them fine-tune on the downstream task.

\section{Results} \label{sec:results}

We first present results on two established PASCAL VOC benchmarks, followed in
\Cref{sec:experiments} by an investigation into different design choices and
pretraining paradigms.

\subsection{PASCAL}

\textbf{VOC 2012 Semantic Segmentation.} We train on the standard extended
segmentation data (10,582 samples) and test on the validation set (1,449
samples). We sample random crops at the original scale. Using our
ResNet-152 model with extended field-of-view we achieve 60.0\% mIU (see
\Cref{tab:comparison}), the
highest reported results on this benchmark that do not use supervised pretraining. It
is noticeable that this value is considerably higher than the AlexNet-based
FCN~\cite{long2015fully} (48.0\%) and even slightly higher than the VGG-16-based FCN
(59.4\%\footnote{Both of these values refer to VOC 2011 and evaluated on only
736 samples, which means the comparison is imprecise.}), both methods trained
on ImageNet.

\textbf{VOC 2007 Classification.} We train on the trainval (5,011 samples) and
test on the test set (4,952 samples). We use the same training procedure with
10-crop testing as in~\cite{donahue2016adversarial}. Our results at 77.3\% mAP
(see \Cref{tab:comparison}) are state-of-the-art when no ImageNet labels are
used.

\section{Experiments} \label{sec:experiments}

\begin{table}
\centering
    \begin{tabular*}{\linewidth}{@{\ab}l|r}
        \toprule
        Pretraining Loss                              & Seg. (\%mIU) \\ \midrule
        Regression\cmt{c95}                           & 48.0\cmt{a159} \\
        Histograms (no hypercolumn)\cmt{c92b}         & 52.7\cmt{a161} \\
        Histograms\cmt{c90e batch size 7, rest are 8} & 52.9\cmt{a160} \\
        \bottomrule
    \end{tabular*}
    \caption{
        \small
        \textbf{Self-supervision loss.} (VGG-16) The choice of loss has a
        significant impact on downstream performance. However, pretraining with
        a hypercolumn does not seem to benefit learning. We evaluate this on
        VOC 2012 Segmentation (val) with a model that uses hypercolumns,
        regardless of whether or not it was used during pretraining.
    }\label{tab:loss}
\end{table}

We present a wide range of experiments, highlighting important aspects of our
competitive results. For these studies, in addition to VOC 2012 Semantic
Segmentation, we also use two classification datasets that we constructed:

\textbf{ImNt-100k/ImNt-10k.} Similar to ImageNet classification with 1000
classes, except we have limited the training data to exactly 100 and 10
samples/class, respectively. In addition, all images are converted to
grayscale. We test on ImageNet val with single center crops of size 224-by-224,
making the results easy to compare with full ImageNet training. For our
pretraining experiments in \Cref{tab:pretraining}, we also use these datasets to
see how well they are able to substitute the entire ImageNet dataset for
representation learning.

\subsection{Loss}
As seen in \Cref{tab:loss}, regressing on color in the L*a*b space yields a
5-point lower result (48.0\%) than predicting histograms in hue/chroma (52.9\%). This 
demonstrates that the choice of loss is of crucial importance to
representation learning. This is a much larger difference than
Larsson~\etal~\cite{larsson2016learning} report in colorization performance
between the two methods ($24.25$ and  $24.45$ dB PSNR / $0.318$ and $0.299$ RMSE).
Histogram predictions are meant to address the problem of color uncertainty.
However, the way they instantiate an image by using summary statistics from the
histogram predictions, means this problem to some extent is re-introduced. Since
we do not care about instantiating images, we do not suffer this penalty and
thus see a much larger improvement using a loss based on histogram predictions.
Our choice of predicting separate histograms in the hue/chroma space also
yields an interesting finding in \Cref{fig:viz}, where we seem to have
non-semantic filters that respond to input with high chromaticity as well as
low chromaticity, clearly catering to the chroma prediction.

\begin{table}
\begin{center}
    \begin{tabular*}{\linewidth}{@{\ab}ll|rrl@{}rr}
        \toprule
        Architecture            & Init.         & Seg.                                    & +FoV                      & ImNt-                                                & 100k                   & 10k \\ 
                                &               & \multicolumn{2}{c}{\footnotesize \%mIU} &                           & \multicolumn{2}{c}{\footnotesize\%top-5} \\ \midrule
        AlexNet                 & Rnd           & 23.5\cmt{a228}                          & 24.6\cup\cmt{a229}            &  & 39.1\cmt{a143c}         & 6.7\cmt{a91 } \\
        AlexNet                 & Col\cmt{c98}  & 36.2\cmt{a219}                          & 40.8\cup\cmt{a227}            &  & 48.2\cmt{a144c}         & 17.4\cmt{a145c} \\ \midrule
        VGG-16                  & Rnd           & 32.8\cmt{a157}                          & 35.1\cup\cmt{a156 or a157}    &  & 43.2\cmt{a90 ,old42.0} & 8.6\cmt{a88 ,old8.0} \\
        VGG-16                  & Col\cmt{c90e} & 50.7\cmt{a178}                          & 52.9\cup\cmt{a160}            &  & 59.0\cmt{a164,old59.0} & 23.3\cmt{a165,old23.0} \\ \midrule
        ResNet-152              & Rnd           & \color{gray}{*9.9\cmt{a186}}            &\color{gray}{*10.5\ceq\cmt{a166}}  &  & 42.5\cmt{a139,old43.0} & 8.1\cmt{a167,old8.0} \\
        ResNet-152              & Col\cmt{c93c} & 52.3\cmt{a177}                          & 53.9\cup\cmt{a168}            &  & 63.1\cmt{a169,old62.0} & 29.6\cmt{a170,old29.0} \\ 
        \bottomrule
    \end{tabular*}
    \caption{
        \small
        \textbf{Architectures.} 
        We compare how various networks perform on downstream tasks on random
        initialization (Rnd) and colorization pretrained (Col). For our
        segmentation results, we also consider the effects of increasing the
        receptive field size (+FoV). Training residuals from scratch (marked
        with a *) is possibly compromised by the hypercolumn, causing the low
        values. 
    }
    \label{tab:archs}
\end{center}
\end{table}

\begin{table}
\begin{center}
    \begin{tabular*}{\linewidth}{@{\ab}lrr|r}
        \toprule
        Pretraining         & Samples      & Epochs & Seg. (\%mIU)          \\ \midrule
        None                & \ndash       & \ndash & 35.1\cmt{a156 or a157}\\ \midrule
        C1000               & 1.3M         & 80     & 66.5\cmt{a135}        \\
        C1000  \cmt{bo53}   & 1.3M         & 20     & 62.0\cmt{a134}        \\
        C1000  \cmt{bo58}   & 100k         & 250    & 57.1\cmt{a125}        \\
        C1000  \cmt{bo59}   & 10k          & 250    & 44.4\cmt{a112}        \\ \midrule
        E10    \cmt{bo63}   & (1.17M) 1.3M & 20     & 61.8\cmt{a232b}       \\
        E50    \cmt{bo64}   & (0.65M) 1.3M & 20     & 59.4\cmt{a235}        \\ \midrule
        H16    \cmt{bo54}   & 1.3M         & 20     & 60.0\cmt{a121}        \\
        H2     \cmt{bo55}   & 1.3M         & 20     & 46.1\cmt{a122}        \\ \midrule
        R50    \cmt{bo56}   & 1.3M         & 20     & 57.3\cmt{a123}        \\
               \cmt{bo56b}  &              & 40     & 59.4\cmt{a172}        \\
        R16    \cmt{bo60b}  & 1.3M         & 20     & 42.6\cmt{a243}        \\ 
               \cmt{bo60}   &              & 40     & 53.5\cmt{a220}        \\
        \bottomrule
    \end{tabular*}

    \includegraphics{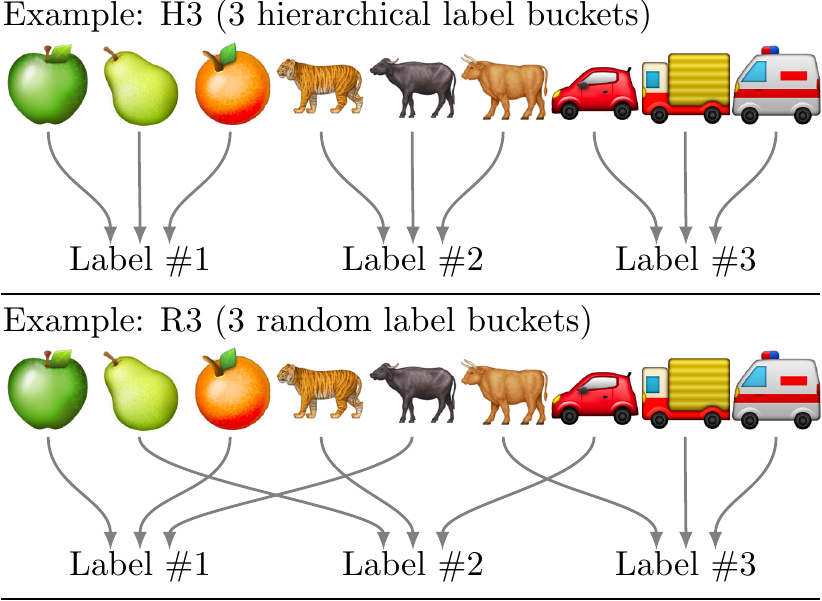}

    \caption{
        \small
        \textbf{ImageNet pretraining.} We evaluate how useful various
        modifications of ImageNet are for VOC 2012 Segmentation (val-gray). We
        create new datasets either by reducing sample size or by reducing the
        label space. The former is done simply by reducing sample size or by introducing 10\% (E10) or 50\% (E50) label noise.
        The latter is done using hierarchical label buckets
        (H16 and H2) or random label buckets (R50 and R16).
        The model trained for 80 epochs
        is the publicly available VGG-16 (trained for 76 epochs) that we
        fine-tuned for grayscale for 4 epochs.  The rest of the models were
        trained from scratch on grayscale images.
    }
    \label{tab:pretraining}
\end{center}
\end{table}

\subsection{Network architecture}
The investigation into the impact of network architecture has been a neglected
aspect of recent self-supervision work, which has focused only on AlexNet.  We
present the first detailed study into the untapped potential of using more
modern networks. These results are presented in \Cref{tab:archs}.

It is not entirely obvious that an increase in model complexity will pay off,
since our focus is small-sample datasets and a smaller network may offer a
regularizing effect.  Take ImNt-100k, where AlexNet, VGG-16, and ResNet-152 all
perform similarly when trained from scratch (39.1\%, 43.2\%, 42.5\%). However,
the percentage point improvement when utilizing colorization pretraining
follows a clear trend (+9.1, +15.8, +20.6).  This shows that self-supervision
allows us to benefit from higher model complexity even in small-sample regimes.
Compare this with $k$-means initialization~\cite{krahenbuhl2016datadriven},
which does not show any improvements when increasing model complexity
(\Cref{tab:comparison}).

Training ResNet from scratch for semantic segmentation is an outlier value in
the table. This is the only experiment that trains a residual network from
scratch together with a hypercolumn; this could be a disruptive combination as
the low numbers suggest.

\subsection{ImageNet pretraining} \label{sec:pretraining}

\begin{table}
\begin{center}
    \begin{tabular*}{\linewidth}{@{\ab}l|rr}
        \toprule
        Initialization & Grayscale input             & Color input             \\ \midrule
        Classification & 66.5 \cmt{a135} & 69.5\cup\cmt{a22 } \\
        Colorization   & 56.0 \cmt{a138} & 55.9\ceq\cmt{a173} \\
        \bottomrule
    \end{tabular*}
    \caption{
        \small
        \textbf{Color vs.\ Grayscale input.} (VOC 2012 Segmentation, \%mIU) Even
        though our classification-based model does 3 points better using color,
        re-introducing color yields no benefit.
    }
    \label{tab:color}
\end{center}
\end{table}

We relate self-supervised pretraining to ImageNet pretraining by revisiting and
reconsidering various aspects of this paradigm (see \Cref{tab:pretraining}).
First of all, we investigate the importance of 1000 classes (C1000).  To do
this, we join ImageNet classes together based on their place in the WordNet
hierarchy, creating two new datasets with 16 classes (H16) and only two classes
(H2). We show that H16 performs only slightly short of C1000 on a downstream
task with 21 classes, while H2 is significantly worse. If we compare this to
our colorization pretraining, it is much better than H2 and only slightly worse
than H16.

Next, we study the impact of sample size, using the subsets ImNt-100k and
ImNt-10k described in \Cref{sec:experiments}.  ImNt-100k does similarly well as
self-supervised colorization (57.1\% vs.\ 56.0\% for VGG-16), suggesting that our method
has roughly replaced 0.1 million labeled samples with 3.7 million unlabeled
samples. Reducing samples to 10 per class sees a bigger drop in downstream
results. This result is similar to H2, which is somewhat surprising: collapsing
the label space to a binary prediction is roughly as bad as using 1/100th of
the training data. Recalling the improvements going from regression to
histogram prediction for colorization, the richness of the label space seems
critical for representation learning.

We take the 1000 ImageNet classes and randomly place them in 50 (R50) or 16 (R16) buckets that we
dub our new labels. This means that we are training a highly complex
decision boundary that may dictate that a golden retriever and a minibus belong
to the same label, but a golden retriever and a border collie do not. We
consider this analogous to self-supervised colorization, since the supervisory
signal similarly considers a red car arbitrarily more similar to a red postbox
than to a blue car. Not surprisingly, our contrived dataset R50 results in a
5-point drop on our downstream task, and R16 even more so with a
20-point drop. However, we noticed that the training loss
was still actively decreasing after 20 epochs. Training instead for 40 epochs
showed an improvement by about 2 points for R50, while 11 points for R16.
In other words, complex classes can provide useful supervision for
representation learning, but training may take longer.
This is consistent with our impression of self-supervised colorization;
although it converges slowly, it keeps improving its feature generality with
more training.

Finally, we test the impact of label noise. When 10\% of the training images
are re-assigned a random label (E10), it has little impact on downstream
performance. Increasing the label noise to 50\% (E50) incurs a 2.6-point penalty, but it is
still able to learn a competitive representation.

\subsection{Training time and learning rate}
\begin{figure}
\begin{center}
    \includegraphics[width=\linewidth,trim=0 0.0cm 0 0]{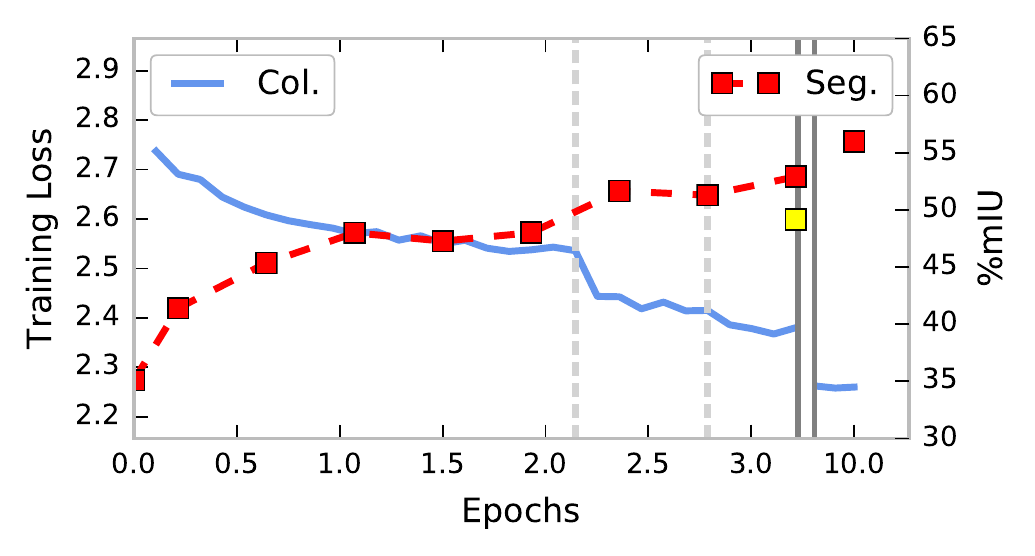}\cmt{a158}
\end{center}
\caption{
\textbf{Learning rate.}
    The blue line shows colorization training loss and the vertical dashed
    lines are scheduled learning rate drops. The red squares are results on a
    downstream task (VOC 2012 Segmentation) initialized by the corresponding
    snapshot of the colorization network. Some key observations: We quickly get
    value for our money, with a 6-point improvement over random initialization
    with only 0.2 epochs of training. Furthermore, improvements on the
    downstream task do not quickly saturate, with results improving further
    when trained 10 epochs in total. Dropping the learning rate on the
    pretraining task helps the downstream task, with a similarly abrupt
    improvement as with the training loss around 2 epochs. Training the full 3
    epochs without ever dropping the learning rate results in 49.1\% (yellow
    square) compared to 52.9\% mIU.
}
\label{fig:droplr}
\end{figure}

We show in \Cref{fig:droplr} that it is crucial for good performance on
downstream tasks to reduce learning rate during pretraining. This result was
not obvious to us, since it is possible that the late stage of training with
low learning rate is too task-specific and will not benefit feature generality.

In addition, we show the importance of training time by demonstrating that
training for three times as long (10 epochs, 37M samples) improves results from
52.9\% to 56.0\% mIU on VOC 2012 Segmentation. Our ResNet-152 model (60.0\%
mIU) trained for 4 months on a single GPU.

\def\n{\tiny\textcolor{black}{$\square$}}
\def\y{\tiny\textcolor{black}{$\blacksquare$}}

\begin{table}
\centering
    \begin{tabular*}{\linewidth}{@{\ab}lr|rrr}
        \toprule
        \multicolumn{2}{l|}{Fine-tuned layers (VGG-16)} & Rnd                           & Col \cmt{Seg (\%mIU)} & Cls                  \\ \midrule
        $\varnothing$   & \n\n\n\n\n\n\n & 3.6\cmt{a153}          & 36.5\cmt{a148} & 60.8\cmt{a147}\\
        {fc6, fc7}      & \n\n\n\n\n\y\y & \dash\cmt{a154}               & 42.6\cmt{a150} & 63.1\cmt{a149}\\
        {conv4\_1..fc7} & \n\n\n\y\y\y\y & \dash\cmt{a155}               & 53.6\cmt{a152} & 64.2\cmt{a151}\\
        {conv1\_1..fc7} & \y\y\y\y\y\y\y & 35.1\cmt{a156: or a157}& 56.0\cmt{a138} & 66.5\cmt{a135}\\
        \bottomrule
    \end{tabular*}
    \caption{
        \small
        \textbf{VOC 2012 Segmentation.} (\%mIU) Classification-based pretraining
        (Cls) needs less fine-tuning than our colorization-based method (Col).
        This is consistent with our findings that our network experiences a
        higher level of feature shift (\Cref{fig:corr-after-finetune}). We
        also include results for a randomly initialized network (Rnd), which
        does not work at all without fine-tuning (3.6\%). This is to show that
        it is not simply by virtue of the hypercolumn that we are able to do
        reasonably well (36.5\%) without any fine-tuning of the base network.
    }
    \label{tab:finetune}
\end{table}

\subsection{Latent representation}
Good results on secondary tasks only give evidence that our self-supervised network has
the potential to be shaped into a useful representation. We investigate if the
representation learned through colorization is immediately
useful or only holds a latent representation. If the latter, how is our
representation different from a good initialization scheme?

\begin{figure*}[t]
    \centering
    \includegraphics[scale=0.84]{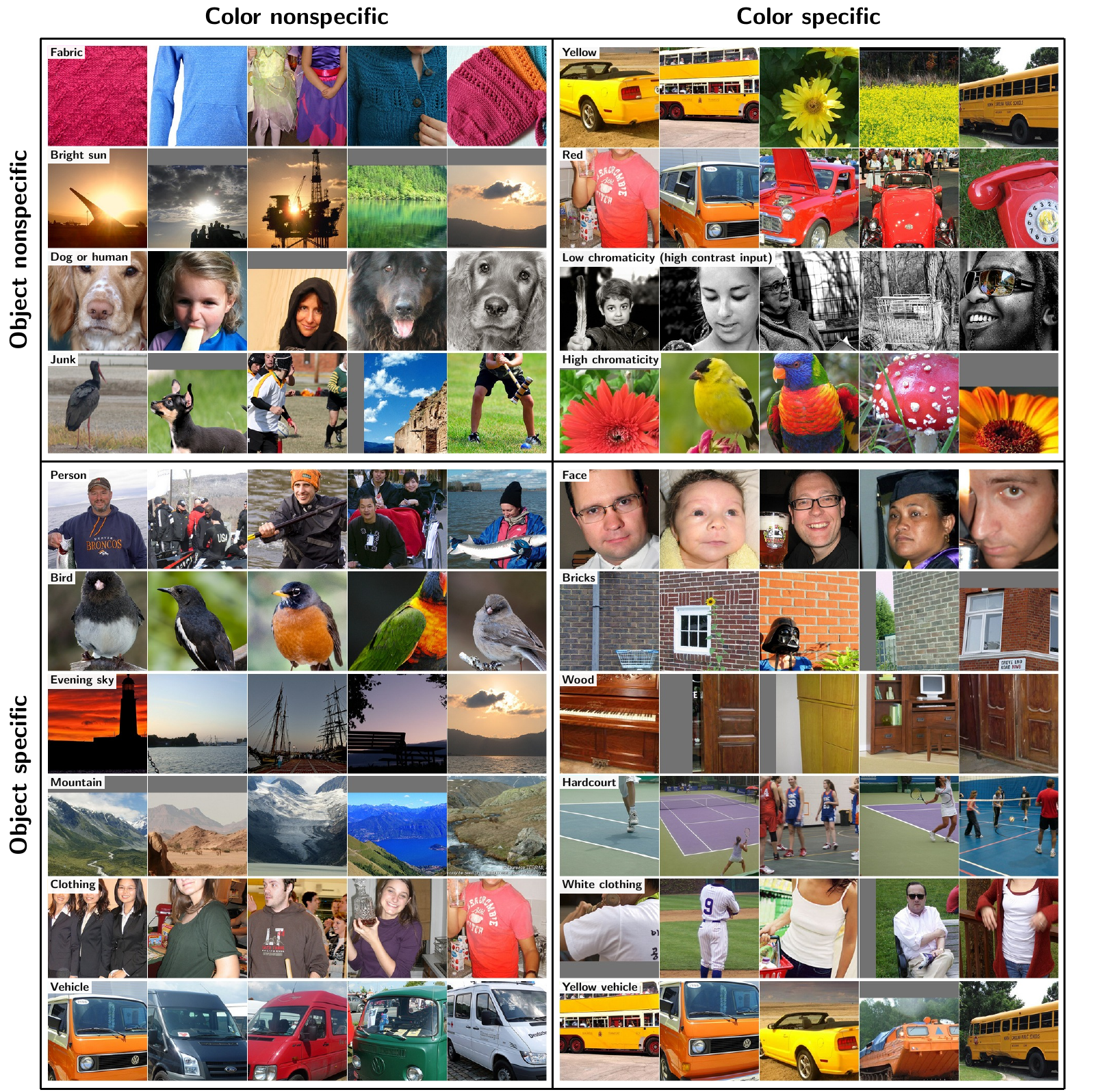}
    \caption{
        \textbf{Feature visualization.}
        Patches around activations from held-out images are visualized for a select number of \texttt{fc7} features (VGG-16). Even though the network takes only grayscale input, we visualize each patch in its original color for the benefit of the reader. As a result, if all the activations are consistent in color (right column), the feature is predictive of color. Similarly, if a feature is semantically coherent (bottom row), it means the feature is predictive of an object class. The names of each feature are manually set based on the top activations.
    }
    \label{fig:viz}
\end{figure*}

First, we visualize features to get a sense of how the colorization network has
organized the input into features. We posit that we will find features
predictive of color, since we know that the colorization network is able to
predict color with good accuracy. In \Cref{fig:viz}, we visualize top
activations from the network's most high-level layer, and indeed we find
color-specific features. However, we also find semantic features that group
high-level objects with great intra-class variation (color, lighting, pose,
etc.).  This is notable, since no labeled data was used to train the network.
The notion of objects has emerged purely through their common color and visual
attributes (compare with~\cite{zhou2015deepscene}). Object-specific features
should have high task generality and be useful for downstream tasks. Features
that are specific to both object and color (bottom-right quadrant in
\Cref{fig:viz}) can be divided into two categories: The first is when the
object generally has a unimodal color distribution (e.g.\ red bricks, brown
wood); the second is when the network has learned a color sub-category of an
object with multimodal color distribution (e.g.\ white clothing, yellow
vehicle). These should all have high task generality, since it is easy for a
task-specific layer to consolidate several color sub-categories into
color-invariant notions of objects.

So how much do the features change when fine-tuned? We visualize top
activations before and after in \Cref{fig:viz-after-finetune} and show
in \Cref{fig:corr-after-finetune} that the colorization features change
much more than label-based features. Some features are completely repurposed,
many are only pivoted, and others remain more or less the same. These
results are consistent with the four quadrants in \Cref{fig:viz}, that show
that some features are specific to colorization, while others seem to have
general purpose.

Next, we look at how much fine-tuning is required for the downstream task.
\Cref{tab:finetune} tells us that even though fine-tuning is more important
than for supervised pretraining (consistent with the correlation results
in \Cref{fig:corr-after-finetune}), it is able to perform the task with the
colorization features alone similarly well as randomly initializing the network
and training it end-to-end from scratch.

Somewhat poor results without fine-tuning and a lower percentage of feature
re-use supports the notion that the colorization network in part holds {\em
latent} features. However, the visualized features and the strong results
overall suggest that we have learned something much more powerful than a good
initialization scheme.

\subsection{Color}
We show in \Cref{tab:color} that re-introducing color yields no
benefit (consistent with the
findings of Zhang~\etal~\cite{zhang2016colorful}). However, concurrent work~\cite{zhang2017split}
presents a better method of leveraging the color channels by separately
training a network for the ``opposite'' task (predicting intensity from color). The
two separate networks are combined for downstream use.

\section{Conclusion}

We have presented a drop-in replacement for ImageNet pretraining, with
state-of-the-art results on semantic segmentation and small-sample
classification that do not use ImageNet labels. A detailed investigation
into self-supervised colorization shows the importance of the loss, network
architecture and training details in achieving competitive results. We draw
parallels between this and ImageNet pretraining, showing that self-supervision
is on par with several methods using annotated data.

\subsection*{Acknowledgments}
We gratefully acknowledge the support of NVIDIA Corporation with the donation
of GPUs used for this research.

{\small
\bibliographystyle{ieee}
\bibliography{arxiv}
}

\appendix

\section{Document changelog}
\label{sec:changelog}
Overview of document revisions:

\appendix
\begin{itemize}
    \item[\textbf{v1}] Initial release.
    \item[\textbf{v2}] CVPR 2017 camera-ready release. Updated ResNet-152
        results in \Cref{tab:comparison} due to additional training for VOC
        2007 Classification ($76.9 \rightarrow 77.3$) and VOC 2012 Segmentation
        ($59.0 \rightarrow 60.0$). Added label noise experiments (E10, E50). Renamed C2/C16 to H2/H16.
        Added references to concurrent work.
    \item[\textbf{v3}] Added overlooked reference.
\end{itemize}

\end{document}